\documentclass[letterpaper, 10 pt, conference]{ieeeconf}  

\IEEEoverridecommandlockouts                              

\usepackage{url}
\usepackage{graphics} 
\usepackage{epsfig} 
\usepackage{mathptmx} 
\usepackage{times} 
\usepackage{amsmath} 
\usepackage{amssymb}  
\usepackage{url}
\usepackage{subfigure}
\usepackage[T1]{fontenc}
\usepackage[vlined,ruled]{algorithm2e}
\usepackage{booktabs}
\usepackage{multirow}

\usepackage{graphicx}
\usepackage{float}

\usepackage{cite}
\usepackage{xcolor}
\usepackage{url}
\usepackage{algpseudocode}
\usepackage[export]{adjustbox}
\usepackage{balance}

\title{\LARGE \bf
	A Robust Laser-Inertial Odometry and Mapping Method for Large-Scale Highway Environments
}

\author{ Shibo Zhao$^{1}$, Zheng Fang$^{1}$, HaoLai Li$^{1}$, Sebastian Scherer$^{2}$
	\thanks{$^{1}$ Shibo Zhao, Zheng Fang and Haolai Li are with Faculty of Robot Science and Engineering, Northeastern University, Shenyang 110819, China; Corresponding author: Zheng Fang, Email:fangzheng@mail.neu.edu.cn}
	\thanks{$^{2}$ Sebastian Scherer is with Robotics Institute, Carnegie Mellon University, Pittsburgh, Pennsylvania 15213 }}

\begin{document}
\maketitle
\thispagestyle{empty}
\pagestyle{empty}

\begin{abstract}
In this paper, we propose a novel laser-inertial odometry and mapping method to achieve real-time, low-drift and robust pose estimation in large-scale highway environments. The proposed method is mainly composed of four sequential modules, namely scan pre-processing module, dynamic object detection module, laser-inertial odometry module and laser mapping module. Scan pre-processing module uses inertial measurements to compensate the motion distortion of each laser scan. Then, the dynamic object detection module is used to detect and remove dynamic objects from each laser scan by applying CNN segmentation network. After obtaining the undistorted point cloud without moving objects, the laser-inertial odometry module uses an Error State Kalman Filter to fuse the data of laser and IMU and output the coarse pose estimation at high frequency. Finally, the laser mapping module performs a fine processing step and the "Frame-to-Model"  scan matching strategy is used to create a static global map. We compare the performance of our method with two state-of-the-art methods, LOAM and SuMa, using KITTI dataset and real highway scene dataset. Experiment results show that our method performs better than the state-of-the-art methods in real highway environments and achieves competitive accuracy on the KITTI dataset. 
\end{abstract}

\section{INTRODUCTION}
\label{sec:introduction}

For intelligent mobile robots, Simultaneous Localization and Mapping (SLAM) is a fundamental technique to achieve the ego-motion estimation and mapping in an unknown environment. Ego-motion estimation is very important for vehicle control, while map is crucial for obstacle perception and path planning.
In recent years, a number of vision-based \cite{barsan2018robust} \cite{runz2017co} \cite{scona2018staticfusion} and laser-based\cite{zhang2018laser} SLAM methods have been proposed, which provides many promising solutions.
 Compared with vision-based SLAM methods, lidar-based SLAM methods can work reliably in visually degraded environments and provide more robust and accurate pose estimation. Nevertheless, lidar-based SLAM algorithms still cannot achieve robust pose estimation in all scenarios. One of the challenging scenarios is highway environments in the field of autonomous driving. 

Highway environments bring many challenges to traditional laser SLAM algorithms. For example:
\begin{itemize}
	\item High speed: Usually, the speed of an autonomous car can at least reach  70 km/h (19.4m/s), which will cause point cloud distortion in each frame and affect the accuracy of mapping.
	
	\item Lack of geometric features: Most of the highway environments lack geometric features in the vertical direction and some places lack forward constraints like a long corridor scene.
	\begin{figure}[t]
		\centering
		\includegraphics[width=0.95\columnwidth]{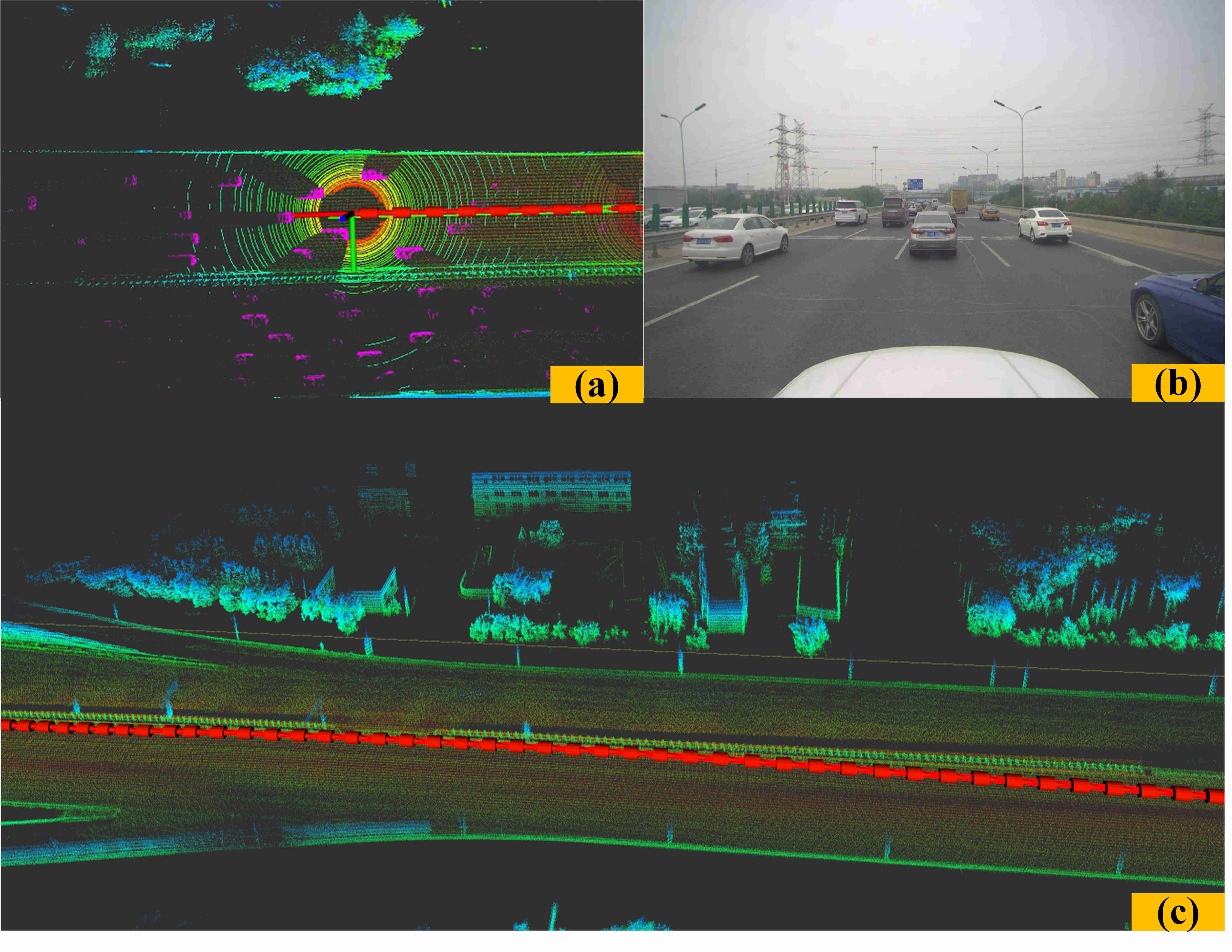}
		\caption{(a) shows the dynamic objects (pink point cloud) are detected from the current laser scan and the motion of sensor (red trajectory) is accurately estimated. (b)  presents the real highway environments corresponding to the current laser scan. We can see that there are many moving vehicles. (c) shows the detail of the highway map built by our method. We can see the windows of the building clearly and the highway road is very clean. }
		\label{fig:iros}
		\vspace{-3mm}
	\end{figure} 
	
	\item Highly dynamic environments: Almost all the time, the current laser scan is negatively affected by dynamic objects.
	\item Less loop closure areas: Since there are less loop closure areas in highway environments, loop closure detection strategy hardly corrects the estimated trajectory.
	
\end{itemize}
Unfortunately, most of the current laser-based SLAM methods assume the surrounding environment is static. Also, since these methods only rely on the measurement provided by the laser scanner, they are hard to deal with aggressive motion in geometrically degraded environments. Because of these two limitations, these methods are difficult to provide robust and accurate motion estimation in the highway scenarios.
 
Motivated by the discussion above, in this work we present a novel  
Laser-Inertial Odometry and Mapping (LIOM) method, which not only achieves robust pose estimation but also builds a static map in highway environments, as shown in Fig \ref{fig:iros}. The main contributions of this paper are as follows:

\begin{itemize}
	\item 
	  We present a real-time lidar odometry and mapping pipeline that utilizes both CNN segmentation network and laser-inertial framework. The CNN segmentation network can remove the influence of dynamic objects in each laser measurement, and the laser-inertial framework can overcome the aggressive motion and achieve robust pose estimation in highway environments.

	\item To verify our proposed solution, extensive experiments are conducted in various challenging environments, such as environments with many moving objects, aggressive motion, and in large scale. Experiment results show that our method can work well in both static and dynamic highway environments.            	
\end{itemize}

The rest of paper is structured as follows: Section \ref{sec:relatedwork} reviews various pose estimation algorithms related to 3D laser scanner. Section \ref{sec:algorithm} describes the details of proposed method. Quantitative and qualitative evaluation on real highway environments and KITTI datasets\cite{Geiger2012CVPR} are presented in Section \ref{sec:experiment}. Section \ref{sec:conclusion} concludes the paper. 
\section {Related Work}
\label{sec:relatedwork}
To enable mobile robots to achieve long-term autonomy in challenging environments, many researchers have been working on how to improve the robustness of laser-based pose estimation methods. Taking motion estimation in highway scenarios as an example, two relevant issues need to be solved: one is how to solve extremely aggressive motion in high-speed driving, and the other one is how to eliminate the influence of dynamic objects and achieve robust pose estimation. 
\subsection{Dealing with Extremely Aggressive Motion }

Due to the low scanning rate of a laser scanner, a scan can not be regarded as a rigid body when the external motion of this sensor is very aggressive. However, most of the laser-based SLAM methods assume the laser points received in a shared "scan frame", which is equivalent to considering that there is no external motion during the sweep. This assumption is not very reasonable and causes what we know as motion distortion.\cite{le20183d}

To solve this problem,  Zlot\cite{bosse2012zebedee} uses the IMU mechanization as the motion model and matches successive laser scan to achieve state estimation.
Le\cite{le20183d} proposes an extrinsic calibration framework, which uses IMU measurements and models the motion distortion of point cloud through pre-integration technique. Ji\cite{zhang2017low} \cite{zhang2018laser} presents a robust multi-sensor fusion pipeline to estimate ego-motion and build a coherent map in various challenging environments. This method utilizes an IMU mechanization for motion prediction and a visual-inertial coupled method to cope with extremely aggressive motion. However, since the above methods are mainly discussed in static scenarios without considering dynamic objects, it will be difficult for these methods to achieve promising results in highway scenes.

\subsection{Dealing with Dynamic Objects}
There have been a large number of promising solutions to SLAM problem in dynamic environments. However, most of these methods are based on vision sensors such as RGB-D camera \cite{runz2017co} \cite{scona2018staticfusion} or stereo camera \cite{barsan2018robust}. In comparison, laser-based methods for dynamic environments are relatively a few. Most of the laser-based SLAM methods assume the surrounding environments are static, which may cause a series of deterioration problems in dynamic environments such as failures in pose estimation and loop closure detection.

To solve this problem, Walcott \cite{walcott2012dynamic} considers the time factor in the map construction process and maintains a precise map in dynamic environments. However, this method is mainly designed for low-dynamic environments, which is not very suitable for highly dynamic environments such as highway scenarios. Fehr \cite{pomerleau2014long} uses a Bayesian model to assess whether a point is  static or dynamic, which can build a coherent map in dynamic environments.  Rendong \cite{wang2019robust} and Masoud \cite{bahraini2018slam} use improved RANSAC algorithm to track moving objects and achieve motion estimation in dynamic environments. However, since the above methods do not integrate IMU measurement, these methods are difficult to deal with aggressive motion and achieve robust pose estimation in highway scene. Jiang \cite{jiang2016static} presents a lidar-camera SLAM system which uses sparse subspace clustering-based motion segmentation method to build a static map in dynamic environments. Nevertheless, the presented system cannot run in real-time and requires sufficient illumination in the environment. Therefore, its application scenarios will be limited. Johannes \cite{graeter2018limo} presents a camera-LiDAR SLAM method which rejects moving objects through semantic labeling. However, The disadvantage of this method is that it only uses lidar to provide depth measurements for visual features and does not utilize lidar to achieve independent pose estimation. Therefore, this method cannot achieve robust pose estimation in visual-degraded environments.  

In recent years, some other new methods are proposed for 3D laser scanner. Jens \cite{behley2018efficient} offers a dense surfel-based approach for motion estimation and mapping. Young-Sik\cite{shin2018direct} presents a camera-LiDAR based SLAM system which is based on direct method and achieves good real-time performance. Jean-Emmanuel \cite{deschaud2018imls} presents a novel laser SLAM method, which uses a specific sampling strategy and new scan-to-model matching. However, since the above methods still assume the surrounding environments are static, it is  difficult for these methods to achieve robust motion estimation in highway environments.

From the discussion in the literature, we can see that the lidar-based pose estimation and mapping in highway environments has not been well solved. This paper aims to achieve real-time, low-drift and robust pose estimation for large-scale highway environments.
   
\section{Laser-Inertial Odometry and Mapping} 
\label{sec:algorithm}
	\begin{figure}[htbp]
		\centering
		\includegraphics[width=0.95\columnwidth]{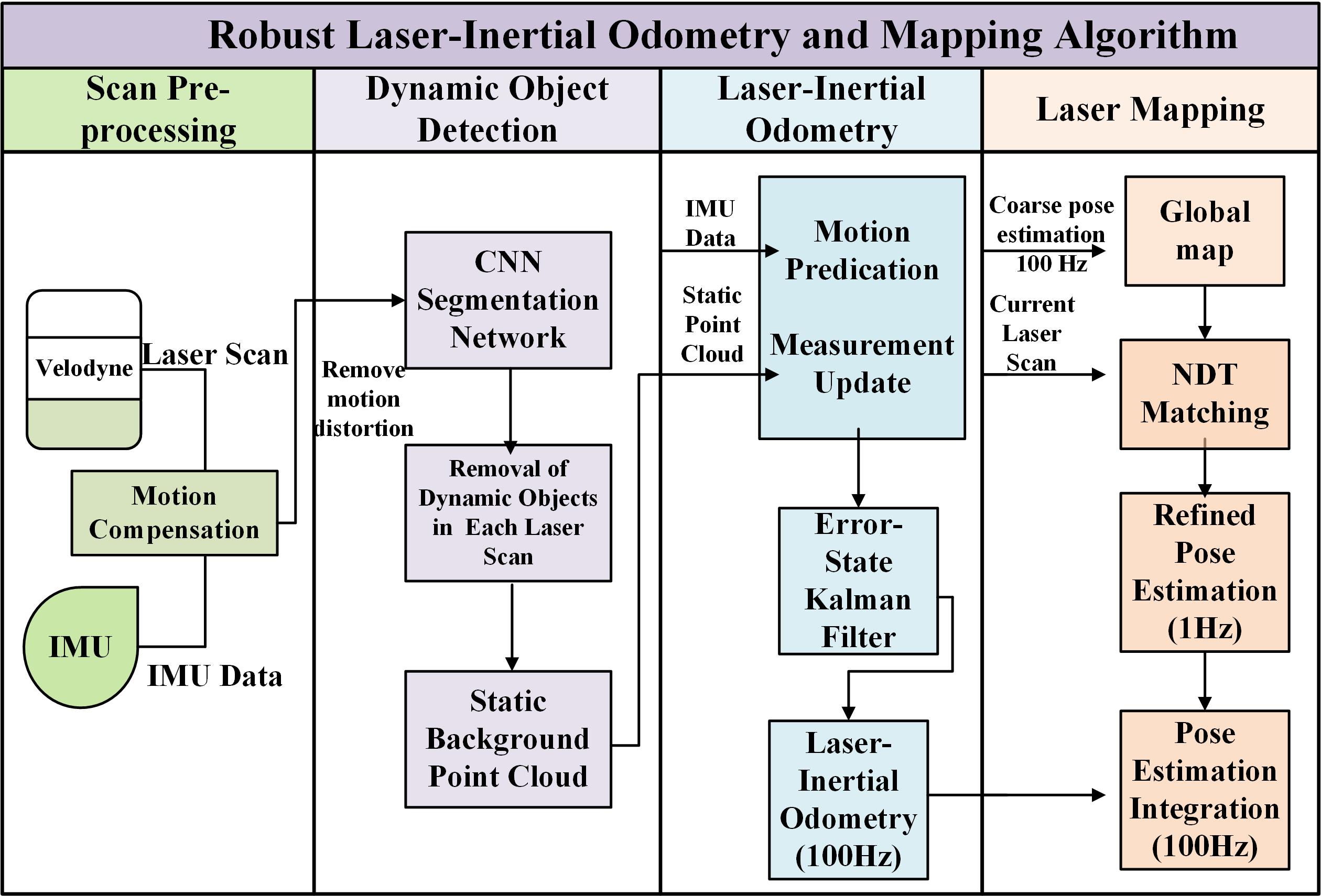}
		\caption{ Overview of Laser-Inertial Odommetry and Mapping Method}
		\label{fig:overview}
		\vspace{-3mm}
	\end{figure} 
In this section, we will introduce the pipeline of our robust Laser-Inertial Odometry and Mapping (LIOM) method as shown in Fig.\ref{fig:overview}, which is composed of four sequential modules namely scan pre-processing module, dynamic object detection module, laser-inertial odometry module and laser mapping module. Cooperation of the above modules allows robust motion estimation and mapping in highway environments.
\subsection{Scan Pre-processing Module}
\begin{figure}[t]
	\centering
	\includegraphics[width=0.7\columnwidth]{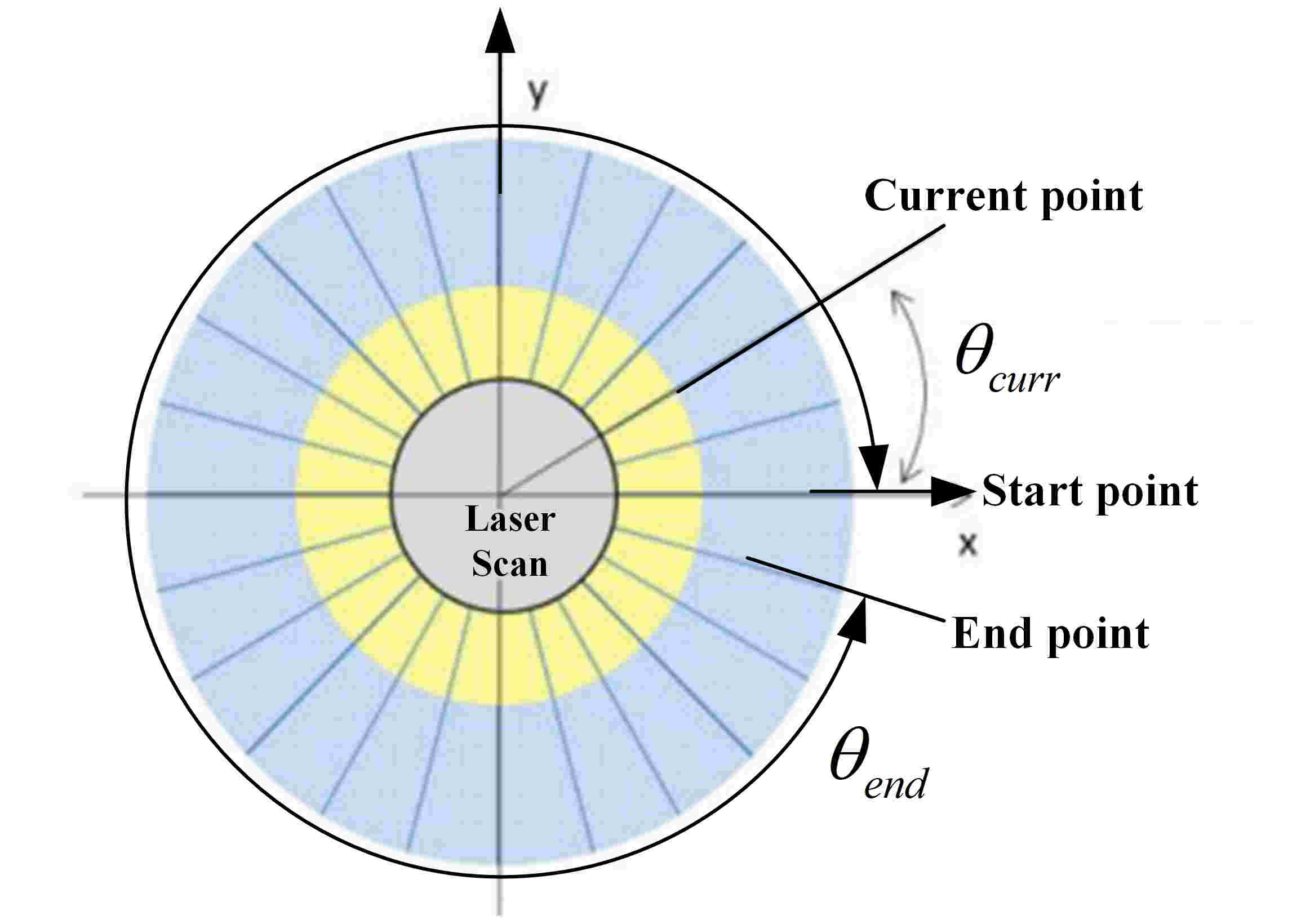}
	\caption{Illustration of numbering timestamp. The start point represents the laser point corresponding to the start scanning time $t_{start}$  of laser scan. Endpoint denotes the laser point corresponding to the end scanning time of laser scan, and its rotation angle relative to the start point is $\theta _{end}$. The current point represents a laser point corresponding to the current scanning time of laser scan, and its rotation angle is 
	 $\theta _{curr}$ relative to the start point. According to Eq.\ref{eq:1}, the corresponding timestamp $t_{curr}$ of current point can be obtained. }
	\label{fig:scanprocess}
	\vspace{-5mm}
\end{figure}

Because of the aggressive motion of the autonomous vehicle in highway environments, point cloud distortion may exist in each laser scan, greatly affecting the accuracy of pose estimation. Therefore, removing the point cloud distortion is very necessary. Here we mainly consider the point cloud motion distortion caused by the non-uniform motion of an autonomous vehicle. The specific implementation steps are as follows:
\begin{itemize}
	\item Step1:  Number all laser points in the laser scan with timestamps as illustrated in Fig. \ref{fig:scanprocess}. According to Eq.\ref{eq:1}, all the laser point in each frame can be numbered with timestamps. 
	
	\begin{equation}
	\label{eq:1}
	{t_{curr}} = {t_{start}} + \partial  \times \frac{{{\theta _{curr}}}}{{{\theta _{end}}}}
	\end{equation} 
	
where $\partial$ represents the scanning period of laser scan. 	
	
	\item Step2: Find the successive IMU measurements with timestamps $k$ and $k+1$, which is the one closest to the timestamp $t_{curr}$ of current laser point.
	
	\item Step3: Through the integral process, we can obtain the pose of IMU in the world coordinate $w$ at time $k$ and $k+1$, $T_k^w = \left[ {{p_x},{p_y},{p_z},{v_x},{v_y},{v_z},{\theta _x},{\theta _y},{\theta _z}} \right] $ and $T_{k + 1}^w = \left[ {{p_x},{p_y},{p_z},{v_x},{v_y},{v_z},{\theta _x},{\theta _y},{\theta _z}} \right]$, respectively.
	
	\item Step4: Using linear interpolation method to determine the pose of IMU in the world coordinate at time $t_{curr}$, which is equivalent to finding the current pose $T_{curr}^w$ of current laser point in the world coordinate. The specific formulas are as follows:
		
		\begin{equation}
		\label{eq:2}
     \left\{ {\begin{array}{*{20}{c}}
     	{rati{o_{front}} = \frac{{{t_{curr}} - {t_{k}}}}{{{t_{k + 1}} - {t_k}}}}\\
     	{rati{o_{back}} = \frac{{{t_{k+1}} - t_{curr}}}{{{t_{k + 1}} - {t_k}}}}\\
     	{T_{curr}^w = T_{k{\rm{ + }}1}^w \times rati{o_{front}} + T_k^w \times rati{o_{back}}}
     	\end{array}} \right.
		\end{equation} 
		
		\item Step5: What we need to solve is the motion distortion of the current laser point relative to the starting point caused by the non-uniform motion. Since the position $P_{curr}^w,P_{start}^w$ and velocity $V_{curr}^w,V_{start}^w$   of the current laser point and start point can be obtained from $T_{curr}^w$  and  $T_{start}^w$, we can get motion distortion $\Delta P_{curr}^{start}$ of laser point in start point coordinate, as shown in Eq. \ref{eq:3}.
		
		\begin{equation}
		\label{eq:3}
	\left\{ {\begin{array}{*{20}{c}}
		{\Delta P_{curr}^w = P_{curr}^w - \left( {P_{start}^w + V_{start}^w \times \partial \frac{{{\theta _{curr}}}}{{{\theta _{end}}}}} \right)}\\
		{\Delta P_{curr}^{start} = R_w^{start}\Delta P_{curr}^w}
		\end{array}} \right.
		\end{equation} 
 
 	\item Step 6: Transform all the laser point in the start point coordinate and subtract their point cloud motion distortion $\Delta P_{curr}^{start}$ for each laser point.

\end{itemize}

\subsection{Dynamic Object Detection Module} 
\label{sec:dection}

In this step, a Fully-Convolutional Neural Networks (FCNN)\cite{ap2} is applied to accurately detect and segment moving objects accurately such as cars, pedestrians and bicycles. The process can be divided into four successive steps: 1. Channel Feature Extraction 2. CNN-Based Obstacle Prediction 3. Obstacle Clustering  4.Post Processing. 

\begin{itemize}
	\item Channel Feature Extraction: In this process, the input point cloud is projected to the 2D X-Y plane. According to the point's X and Y coordinates, each point within a preset range is quantized into one cell of the 2D grid. Then, 8 statistical measurements of the points for each cell are computed and fed into the FCNN. We refer readers to \cite{ap2} for more details on 8 statistical measurements.
	
	\item CNN-Based Obstacle Prediction: FCNN is used to predict the cell-wise obstacle attributes including the centre offset, objectness, positiveness, object height and class probability. 
	
	\item Obstacle Clustering: Above five cell object attributes are used to generate obstacle objects. Then,  a compressed union-find algorithm is used to find candidates of obstacle cluster. 
	
	\item Post-Processing: After obstacle clustering, we obtain a set of candidate object clusters. Post-processing further refines potential candidate clusters and outputs the final clusters by predesigned parameters.	
\end{itemize}

These four successive steps will remove dynamic objects in highway environments, and the remaining static background clouds will be used for pose estimation and mapping.

\subsection{Laser-Inertial Odometry Module} 
For robust pose estimation in highway scenes, we use Error State Kalman Filter (ESKF) to realize sensor fusion between laser scanner and IMU. Compared with the traditional Kalman filtering method, this method mainly has two advantages. First of all, the error-states are small, which means that
we can omit all second-order products and reduce computation. Second, the orientation error-state is minimal, which avoids the over-parameterization  and gimbal lock issues.

In this section, we will give a brief description of the motion prediction and measurement update. The motion prediction is used to provide state prediction through IMU measurements, while the measurement update is used to correct predicated states. The more detail can be found in\cite{Zhen-2017-18126} \cite{sola2017quaternion}.

\subsubsection{Motion Prediction}
The motion prediction is composed of 2 parts: Error-state dynamics and Propagation process.

\paragraph{Error-state dynamics}
According to previous work\cite{sola2017quaternion}, the relation of true state $x$, predicated state $\hat x$ and error state $\delta x$ can be described in Eq.\ref{eq:4}.

\begin{equation}
\label{eq:4}
x = \hat x \oplus \delta x
\end{equation} 
Here $\oplus $ represents a generic composition. 

The error-state dynamic systems are listed as follows:



\begin{equation}
\label{eq:6}
\delta x = \left[ {\begin{array}{*{20}{c}}
	{\delta \dot v}\\
	{\delta \dot p}\\
	{\delta \dot \theta }\\
	{\delta {{\dot a}_b}}\\
	{\delta {{\dot w}_b}}
	\end{array}} \right] = \left[ {\begin{array}{*{20}{c}}
	{ - \hat R{{\left[ {{a_m} - {{\hat a}_b}} \right]}_ \times }\delta \theta  - \hat R\delta {a_b} - \hat R{a_n}}\\
	{\delta v}\\
	{ - {{\left[ {{w_m} - {{\hat w}_b}} \right]}_ \times }\delta \theta  - \delta {w_b} - {w_n}}\\
	{{a_w}}\\
	{{w_w}}
	\end{array}} \right]
\end{equation} 	
where $v$ and $p$ represent the velocity and position respectively in the global frame. $R$ denotes the rotation matrix. $a_b$ and $w_b$ are accelerometer and gyroscope bias.  ${a_n}$ and ${w_n}$  represent accelerometer and gyroscope noise, ${a_m}$ and ${w_m}$  are the acceleration and angular velocity measurements. ${a_w}$ and ${w_w}$ are the Gaussian random walk noise of biases. $\delta \theta$ is represented by a $3 \times 1 $  angle vector.
\paragraph{Propagation process} The propagation process contains predicated state propagation and error covariance propagation. We directly use Euler integration to propagate the predicated state. The error covariance matrix is derived from linearizing the error state kinematics. The expression of propagation rule is described in Eq.\ref{eq:7} and Eq.\ref{eq:8}.

\begin{equation}
\label{eq:7}
\left[ {\begin{array}{*{20}{c}}
	{{{\hat v}_{t + 1}}}\\
	{{{\hat p}_{t + 1}}}\\
	{{{\hat R}_{t + 1}}}\\
	{{{\hat a}_{b\left( {t + 1} \right)}}}\\
	{{{\hat w}_{b\left( {t + 1} \right)}}}
	\end{array}} \right] = \left[ {\begin{array}{*{20}{c}}
	{{{\hat v}_t} + \left[ {{{\hat R}_t}\left( {{a_m} - {{\hat a}_{b\left( t \right)}}} \right) + g} \right]\Delta t}\\
	{{{\hat p}_t} + {{\hat v}_t}\Delta t + \frac{1}{2}\left[ {{{\hat R}_t}\left( {{a_m} - {{\hat a}_{b(t)}}} \right) + g} \right]\Delta {t^2}}\\
	{{{\hat R}_t}R\left\{ {\left( {{w_m} - {{\hat w}_{b(t)}}} \right)\Delta t} \right\}}\\
	{{{\hat a}_{b\left( t \right)}}}\\
	{{{\hat w}_{b\left( t \right)}}}
	\end{array}} \right]
\end{equation} 
\vspace{-3mm}
\begin{equation}
\label{eq:8}
{\hat \Sigma _{t + 1}} = {F_x}{\Sigma _t}{F_x}^T + {F_n}{Q_n}{F_n}^T
\end{equation} 
where ${F_x}$ and ${F_n}$ are the Jacobian matrix of Eq.\ref{eq:7} respect to error state and noise vector; ${Q_n}$ is the covariance matrix of noise vector. The more detail can be found in \cite{sola2017quaternion}.
%
%

\subsubsection{Measurement Update}
The measurement update is composed of 4 parts: Observation model, Recover measurement, Correction and Reset nominal states.

\paragraph{Observation model} 
In error state format, the observation model can be written as
\begin{equation}
\label{eq:9}
\delta y = H\delta x = \left[ {\begin{array}{*{20}{c}}
	0&{{I_3}}&0&0&0\\
	0&0&{{I_3}}&0&0
	\end{array}} \right]\delta x
\end{equation} 

\paragraph{Recover Measurement } 
After the motion prediction process, we obtain the pose prior $ \delta \bar x_{t + 1}^m \in \mathbb{R} {^6}$, $\bar \Sigma _{t + 1}^m \in \mathbb{R} {^{6 \times 6}}$ and use them as an initial guess for the multi-threaded Normal Distributions Transform (NDT) method. Then, the pose posterior $\delta x_{t + 1}^m \in \mathbb{R} {^6}$ and $\Sigma _{t + 1}^m \in \mathbb{R}{^{6 \times6}}$ are calculated from scan matching. Finally, the measurement $\delta y_{t + 1}^m \in \mathbb{R} {^6}$  and noise $C_{t + 1}^m$  are recovered by inversing the KF measurement update.	
\begin{equation}
\label{eq:10}
C_{t + 1}^m = {\left( {\Sigma {{_{t + 1}^m}^{ - 1}} - {{\left( {\bar \Sigma _{t + 1}^m} \right)}^{ - 1}}} \right)^{ - 1}}
\end{equation} 

\begin{equation}
\label{eq:11}
\delta y_{t + 1}^m = {\left( {{K^m}} \right)^{ - 1}}\left( {\delta x_{t + 1}^m - \delta \bar x_{t + 1}^m} \right) + \delta \bar x_{t + 1}^m
\end{equation} 
where ${K^m} = {\bar \Sigma _{t + 1}}{H^{mT}}{\left( {{H^m}{{\bar \Sigma }_{t + 1}}{H^{mT}} + C_{t + 1}^m} \right)^{ - 1}} \in \mathbb{R} {^{6 \times 6}}
$  is the Kalman gain. 

\paragraph{Correction} 
Once the measurements $\delta y_{t + 1}^m \in \mathbb{R} {^6}$ are calculated, the full error state posterior $\delta x_{t + 1}^m \in \mathbb{R} {^6}$ and covariance ${\Sigma _{t + 1}} \in \mathbb{R}  {^{15 \times 15}}$ can be updated through Eq.(\ref{eq:12}-\ref{eq:14}). The $ K \in \mathbb{R}{^{15 \times 6}}$ represents Kalman gain of the normal KF update.

\begin{equation}
\label{eq:12}
K = {\bar \Sigma _{t + 1}}{H^T}\left( {H{{\bar \Sigma }_{t + 1}}{H^T} + {C_{t + 1}}} \right)
\end{equation} 
\begin{equation}
\label{eq:13}
\delta {x_{t + 1}} = K\left( {\delta {y_{t + 1}} - H\delta {{\bar x}_{t + 1}}} \right)
\end{equation} 
\begin{equation}
\label{eq:14}
{\Sigma _{t + 1}} = \left( {{I_{15}} - KH} \right){\bar \Sigma _{t + 1}}
\end{equation} 
\paragraph{Reset Nominal States} 

Finally, the nominal state get updated through Eq.\ref{eq:15}.   
      
\begin{equation}
\label{eq:15}
\left[ {\begin{array}{*{20}{c}}
	{{{\hat v}_{t + 1}}}\\
	{{{\hat p}_{t + 1}}}\\
	{{{\hat R}_{t + 1}}}\\
	{{{\hat a}_{b\left( {t + 1} \right)}}}\\
	{{{\hat w}_{b(t + 1)}}}
	\end{array}} \right] = \left[ {\begin{array}{*{20}{c}}
	{{{\hat v}_t} + \delta {v_{t + 1}}}\\
	{{{\hat p}_t} + \delta {p_{t + 1}}}\\
	{{{\hat R}_t} \cdot R\left( {\delta {\theta _{t + 1}}} \right)}\\
	{{{\hat a}_{bt}} + \delta {a_{b\left( {t + 1} \right)}}}\\
	{{{\hat w}_{bt}} + \delta {w_{b\left( {t + 1} \right)}}}
	\end{array}} \right]
\end{equation}
After completing the above processes, robust laser-inertial odometry at high frequency (100Hz) is obtained. 

\subsection{Laser Mapping Module}  
 For further improving the accuracy of pose estimation, we adopt the similar mapping strategy of LOAM \cite{zhang2017low} algorithm. The main difference between ours and LOAM algorithm is that we apply multi-threaded NDT algorithm \cite{magnusson2007scan}\footnote{\url{https://github.com/koide3/ndt_omp}} to achieve "Frame-to-Model" scan matching rather than a feature-based method. The reason is that the NDT based method is more robust compared with a featured based method. Therefore, it can provide more accurate motion estimation in geometrically degraded environments, especially for highway scenes.
 
\section{Experiments and Results} 
\label{sec:experiment}
In this section, we will compare the performance of our method with two state-of-the-art methods, LOAM \cite{zhang2017low} and SuMa\cite{behley2018efficient}\footnote{\url{https://github.com/jbehley/SuMa}}, in various challenging environments, such as environments with dynamic objects, aggressive motion, and  in large scale. The whole set of experiments can be divided into comparison of pose estimation in real highway environments and comparison of pose estimation on KITTI datasets\cite{Geiger2012CVPR}. Our experiment video is available at: \textcolor{blue}{\url{https://youtu.be/AQJ9wFK8jwQ}}  

\subsection{The comparison of pose estimation in real highway environments} 
To evaluate the performance of different methods in real highway environments, we select datasets collected by Tencent Autonomous Driving Lab. The datasets contain various challenges such as high speed (with maximum linear and angular speed are $72km/h$ and ${60^ \circ }/s$ respectively), insufficient geometric features, high-dynamic and less loop closure areas. We carried out six experiments on this dataset. According to different scenarios, these six groups of experiments can be divided into two categories: (1) Accuracy comparison of pose estimation in static highway environments (static 06 dataset); (2) Robustness comparison of pose estimation in dynamic highway environments (dynamic 01-05 dataset);

\begin{figure}[h]
	\subfigure[static highway scene]{
		\label{scene1}
		\includegraphics[width=0.45\columnwidth]{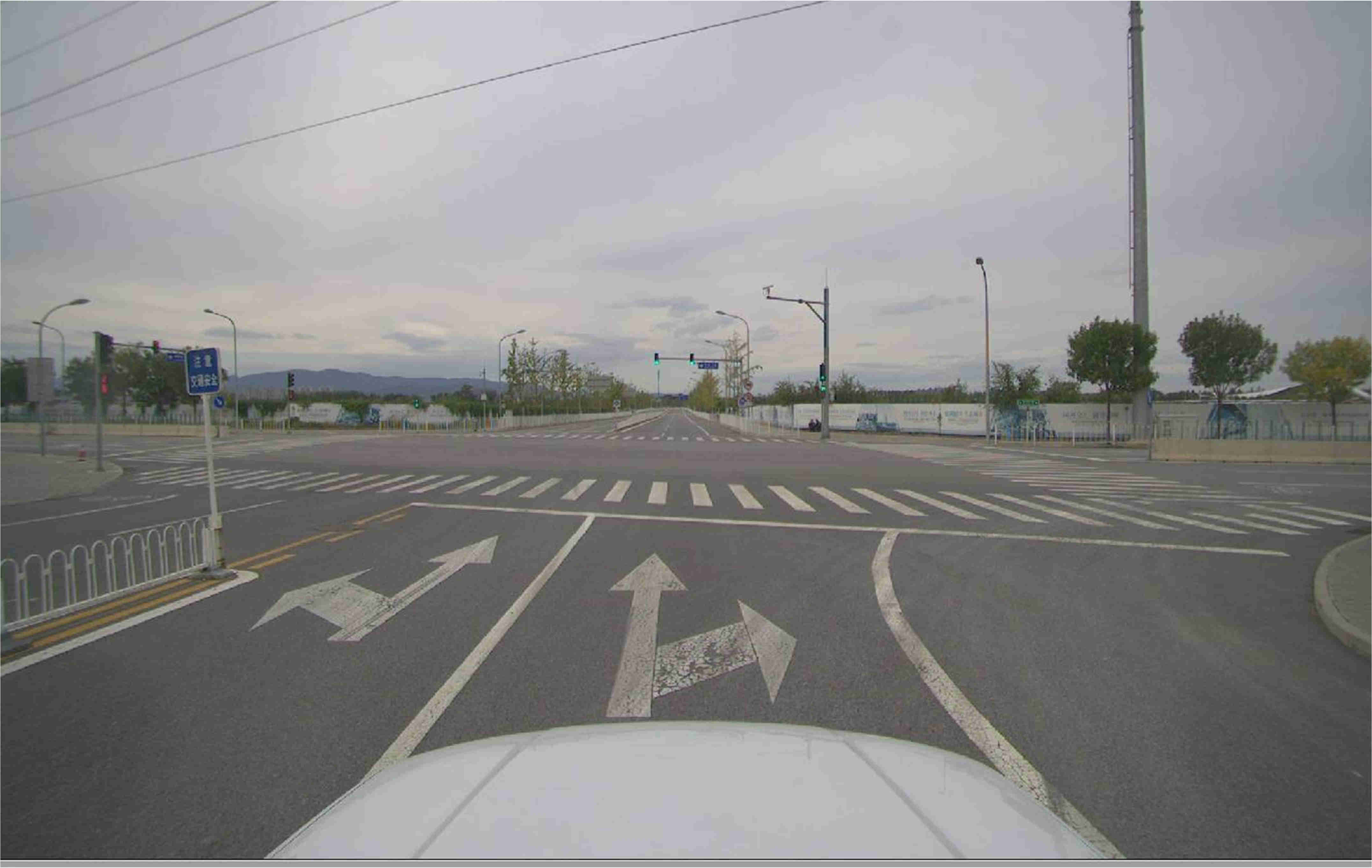}
	}
	\subfigure[dynamic highway scene]{
		\label{scene2}
		\includegraphics[width=0.48\columnwidth]{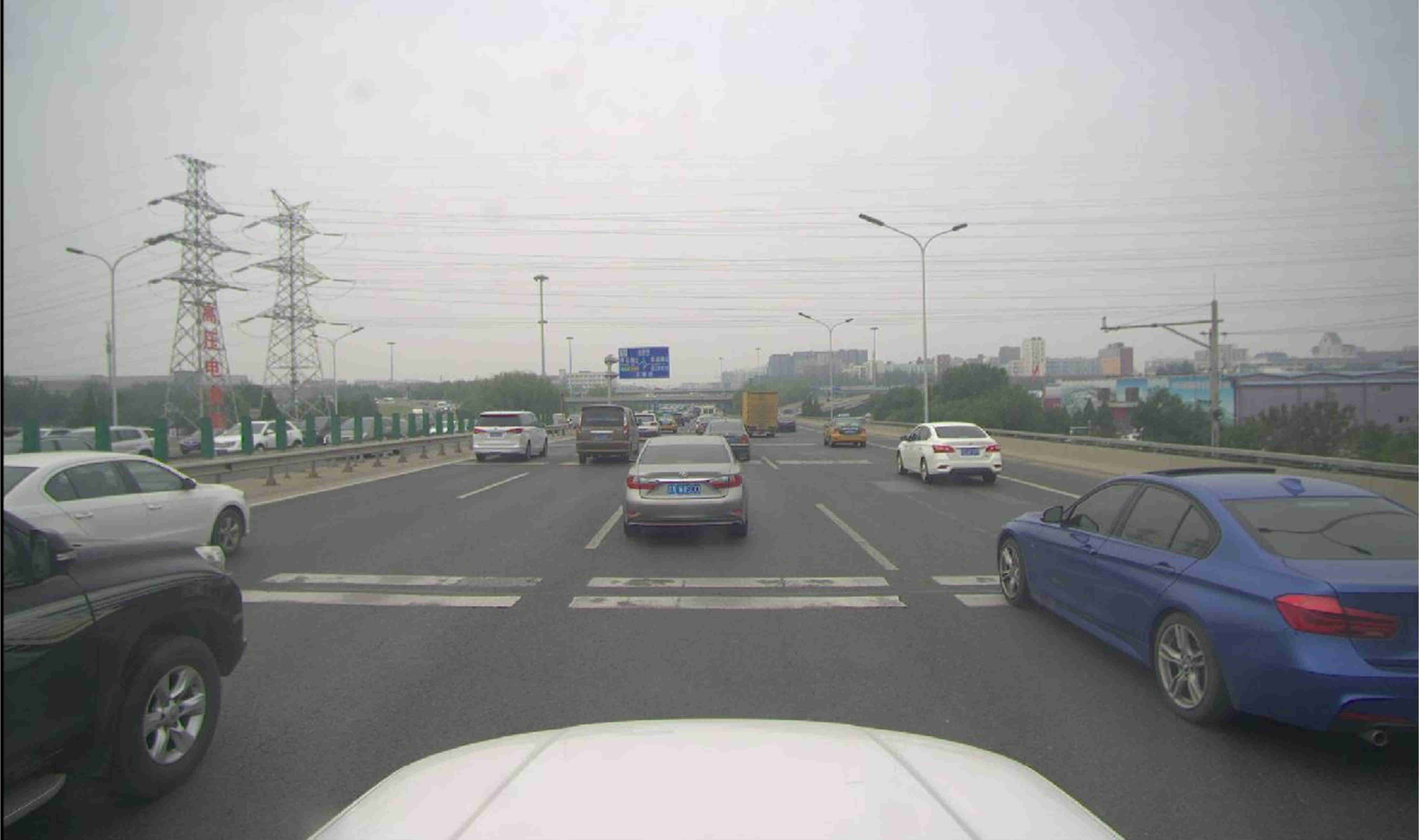}
	}
	\vspace{-5mm}
	\caption{The real scene of highway}
	\label{fig:realscene}
	\vspace{-5mm}
\end{figure}

\subsubsection{Accuracy comparison of pose estimation in static highway environments} We first test the performance of ours, LOAM and SuMa algorithms on the static 06 dataset. A snapshot of the real scene is shown in Fig.\ref{fig:realscene}(a). We can see that the highway environment does not have sufficient structure information in forward and vertical direction. This will result in insufficient geometric features and constraints in these directions for each received laser scan. After 6.3 km of driving within 7 min, a highway map is built. The maximum speed over the test is 16.7m/s and the maximum angular speed exceeds ${60^ \circ }/s $. 

\begin{figure*}[h]
	\subfigure[OURS]{
		\label{ours}
		\includegraphics[width=0.65\columnwidth]{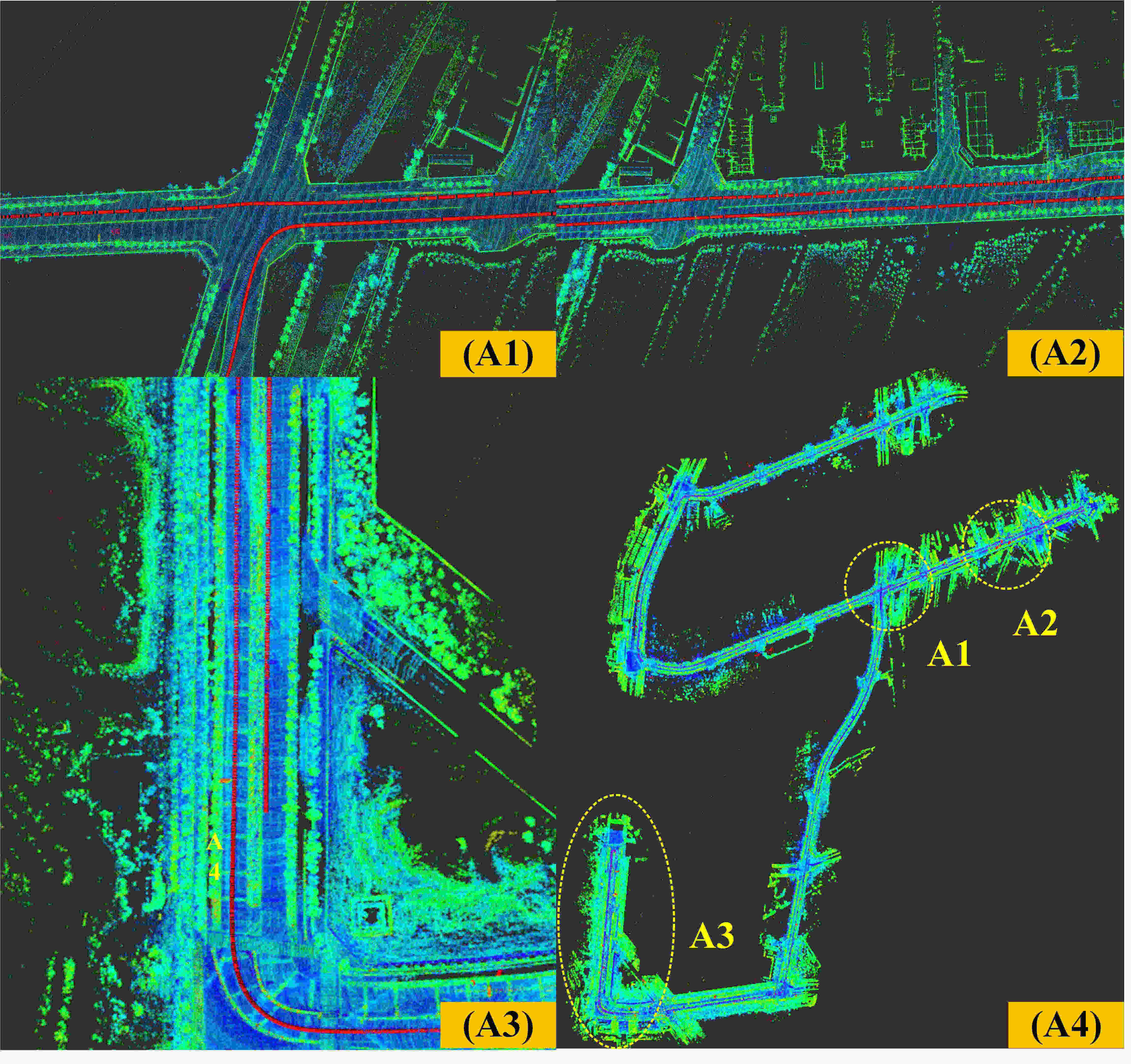}
	}
	\hspace{0.0001\columnwidth}
	\subfigure[SuMa]{
		\label{suma}
		\includegraphics[width=0.62\columnwidth]{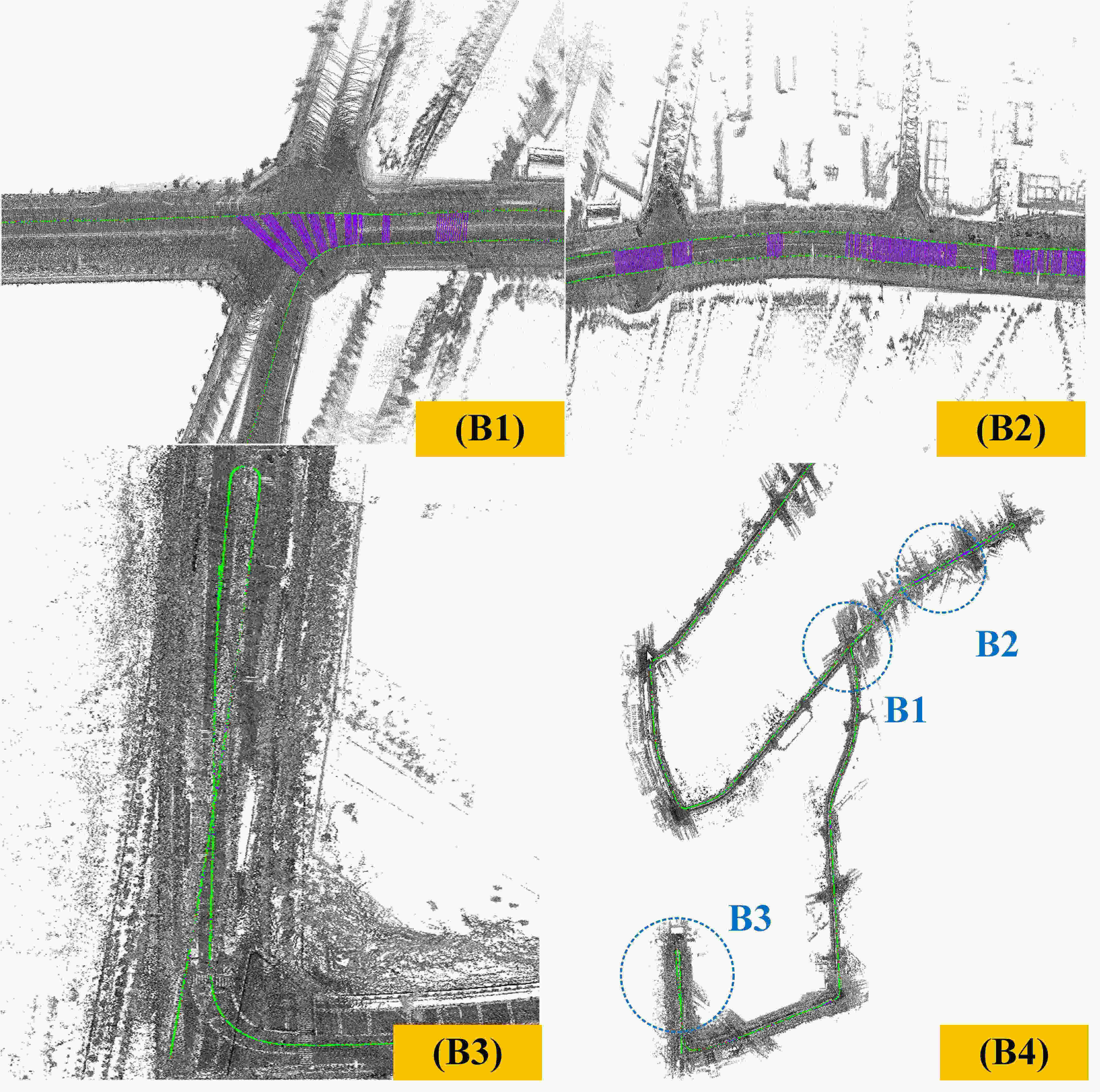}
	}
	\hspace{0.0001\columnwidth}
	\subfigure[LOAM]{
		\label{loam}
		\includegraphics[width=0.67\columnwidth]{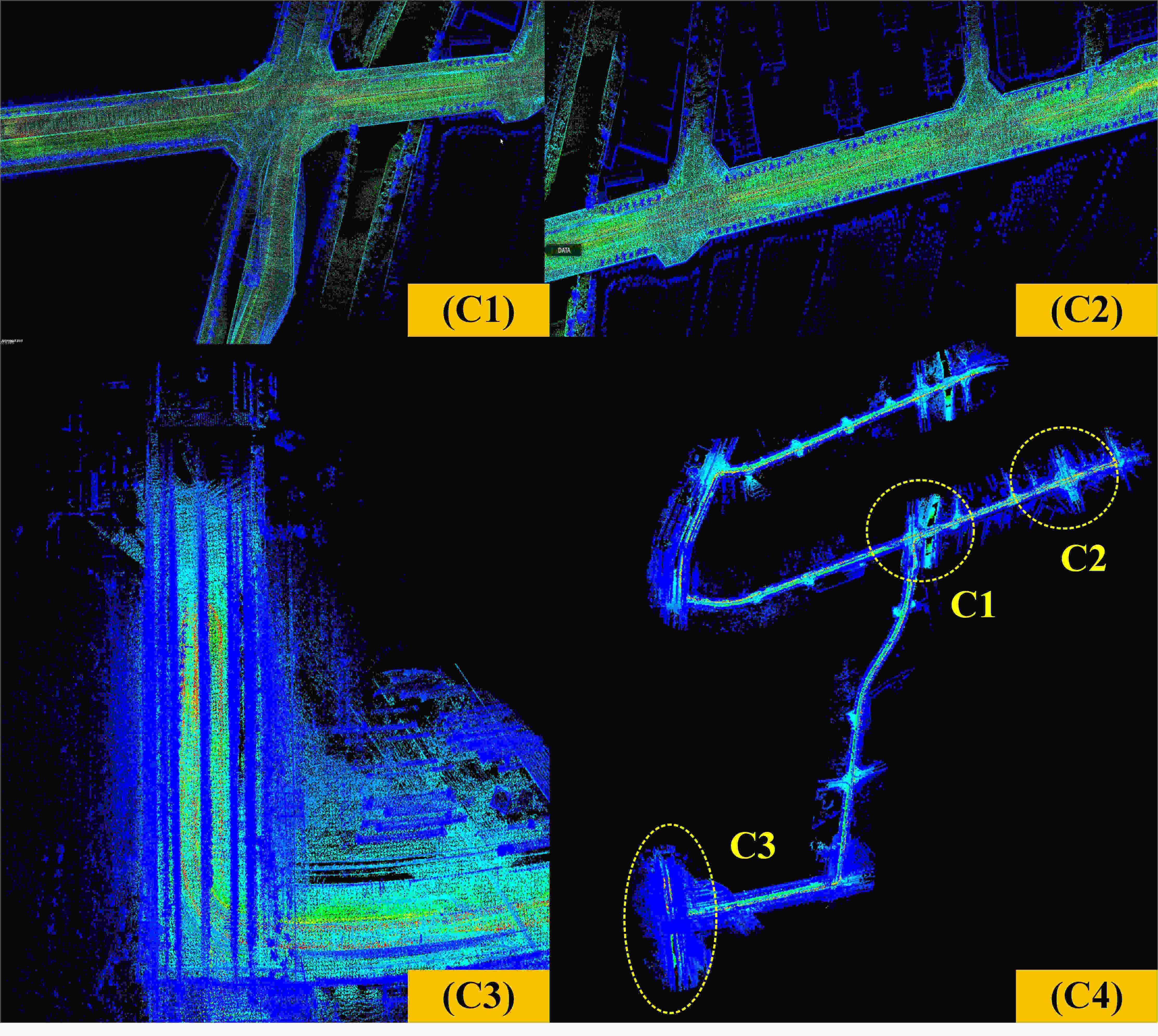}
	}
	
	\caption{The comparison of mapping after applying Ours, SuMa and LOAM algorithms respectively in static highway environments. (A1-A3),(B1-B3) and (C1-C3) present three zoom-in views respectively on loop closure areas for readers to compare the local registration accuracy of different methods.}
	\label{fig:static}
\end{figure*}

To evaluate the accuracy of pose estimation, we mainly compare the detail of loop closure areas and the quality of global map. The results are shown in Fig.\ref{fig:static}(a), Fig.\ref{fig:static}(b) and Fig.\ref{fig:static}(c) respectively. In Fig.\ref{fig:static}(A4), we show the global map built by our method and present three zoom-in views (Fig.\ref{fig:static} (A1-A3)) of loop closure areas (A1-A3) for inspecting the local registration accuracy. The red line in Fig.\ref{fig:static}(A1-A3) shows the estimated trajectory of our method. Carefully examining these figures, we could find that the detail of the loop closure area is very clear, which means that the estimated trajectory is very accurate. This is because the proposed method makes full use of the prior angular and translation information provided by IMU in a short time. Therefore, the current point cloud can obtain a good initial guess and match the global map accurately.

In contrast, the global map corresponding to LOAM algorithm (Fig.\ref{fig:static}(C4)) has large point cloud distortion, especially in the C3 area. This is because LOAM algorithm relies heavily on the geometric features of laser range data and 
C3 area is exactly lack of geometric features in forward direction. Therefore, this makes it difficult for LOAM algorithm to build correct geometric constraints and leads the failure of pose estimation. The performance of SuMa algorithm is also not satisfying and its corresponding map and details are shown in Fig.\ref{fig:static}(B1-B4). The reason is similar to LOAM algorithm.

\subsubsection{Robustness comparison of pose estimation in dynamic highway environments}
We further test the performance of ours, LOAM and SuMa algorithms on dynamic 05 dataset. The vehicle is driven from highway to urban areas for 4.58 km of travel and its speed is between 13.7m/s and 20 m/s during the test. Each frame of laser scan is affected by a large number of moving vehicles as shown in Fig.\ref{fig:realscene}(b).
  
\begin{figure*}[h]
	
	\subfigure[OURS]{
		\label{ours2}
		\includegraphics[width=0.67\columnwidth]{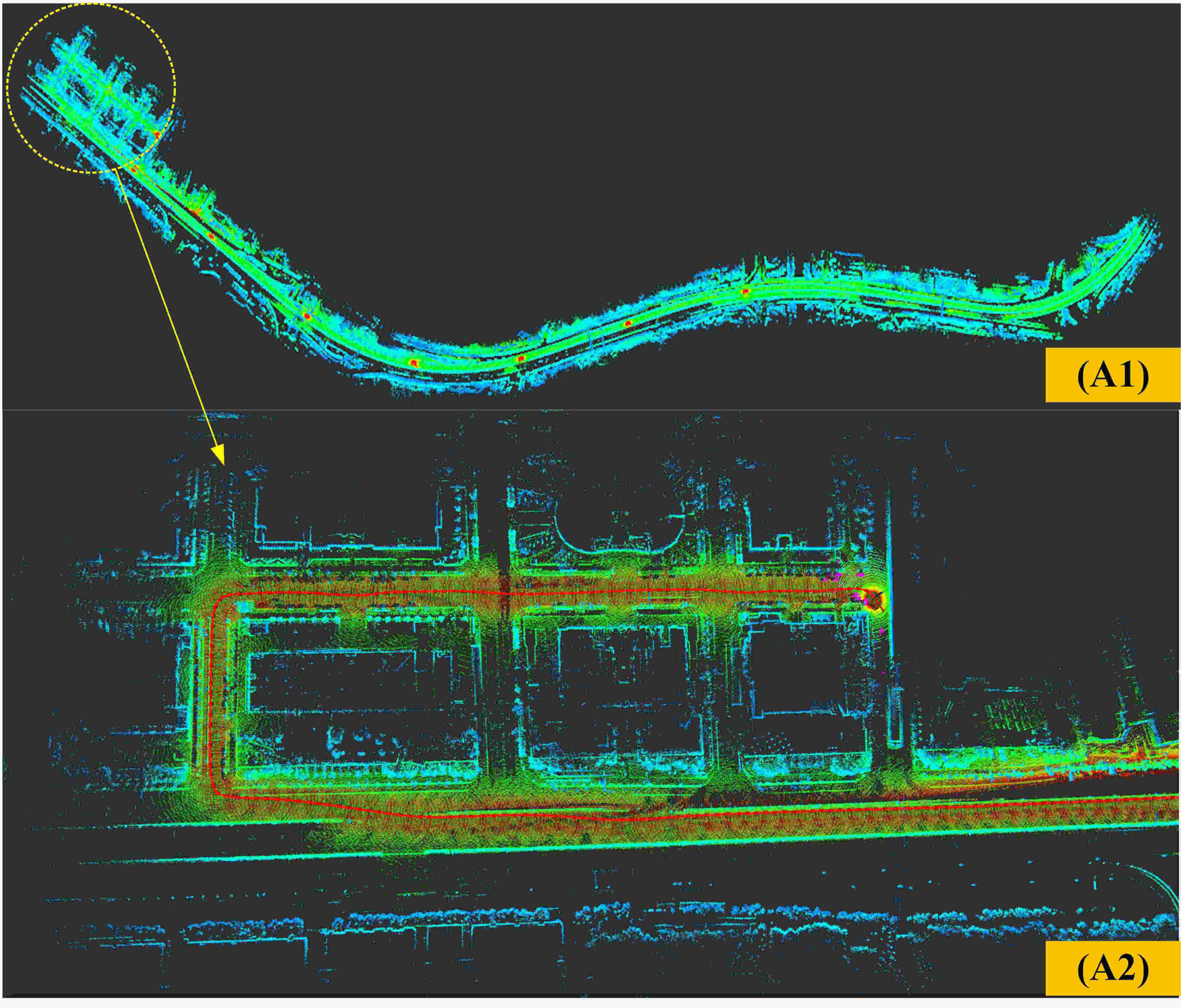}
	}
	\hspace{0.000001\columnwidth}
	\subfigure[SuMa]{
		\label{suma2}
		\includegraphics[width=0.6\columnwidth]{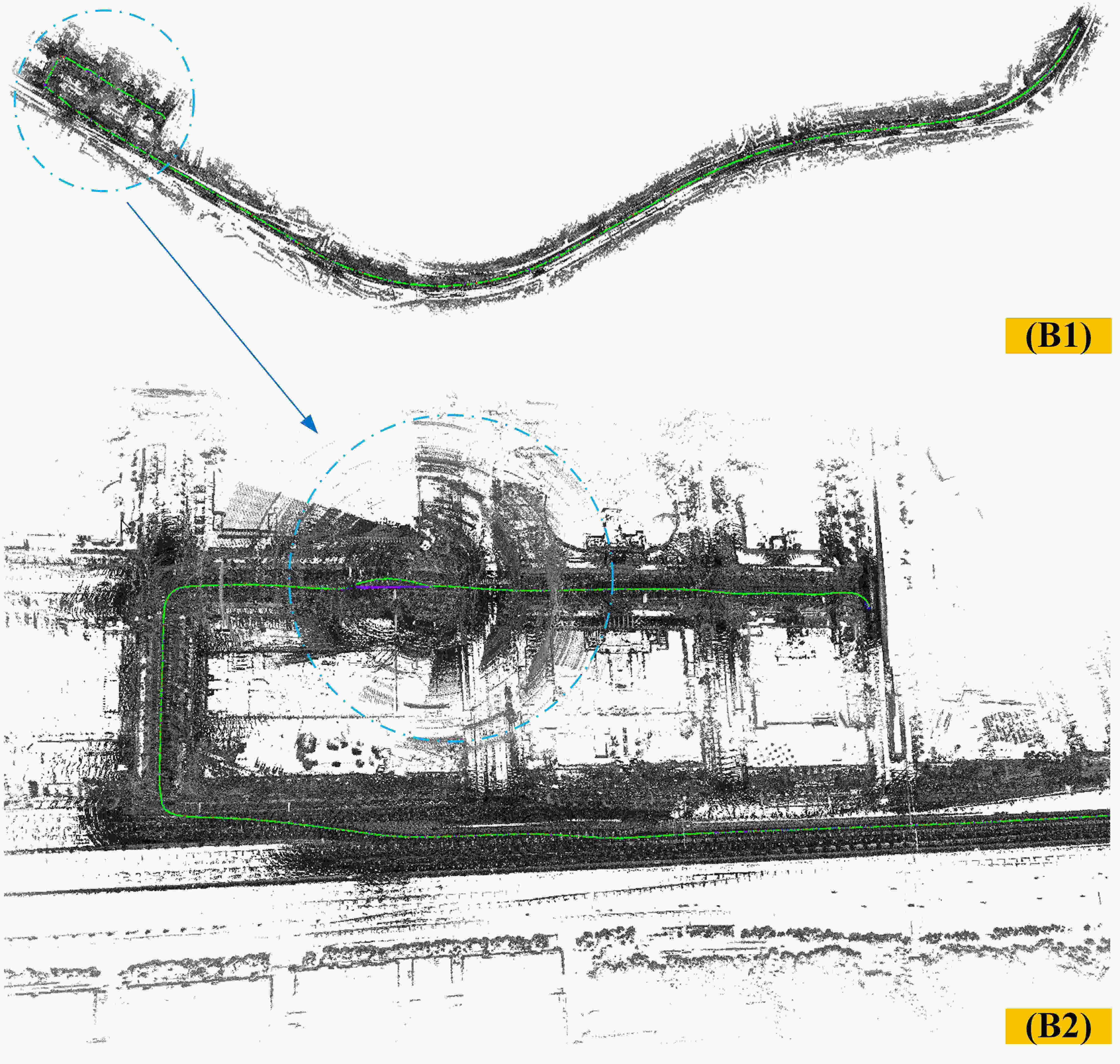}
	}
	\hspace{0.000001\columnwidth}
	\subfigure[LOAM]{
		\label{loam2}
		\includegraphics[width=0.62\columnwidth]{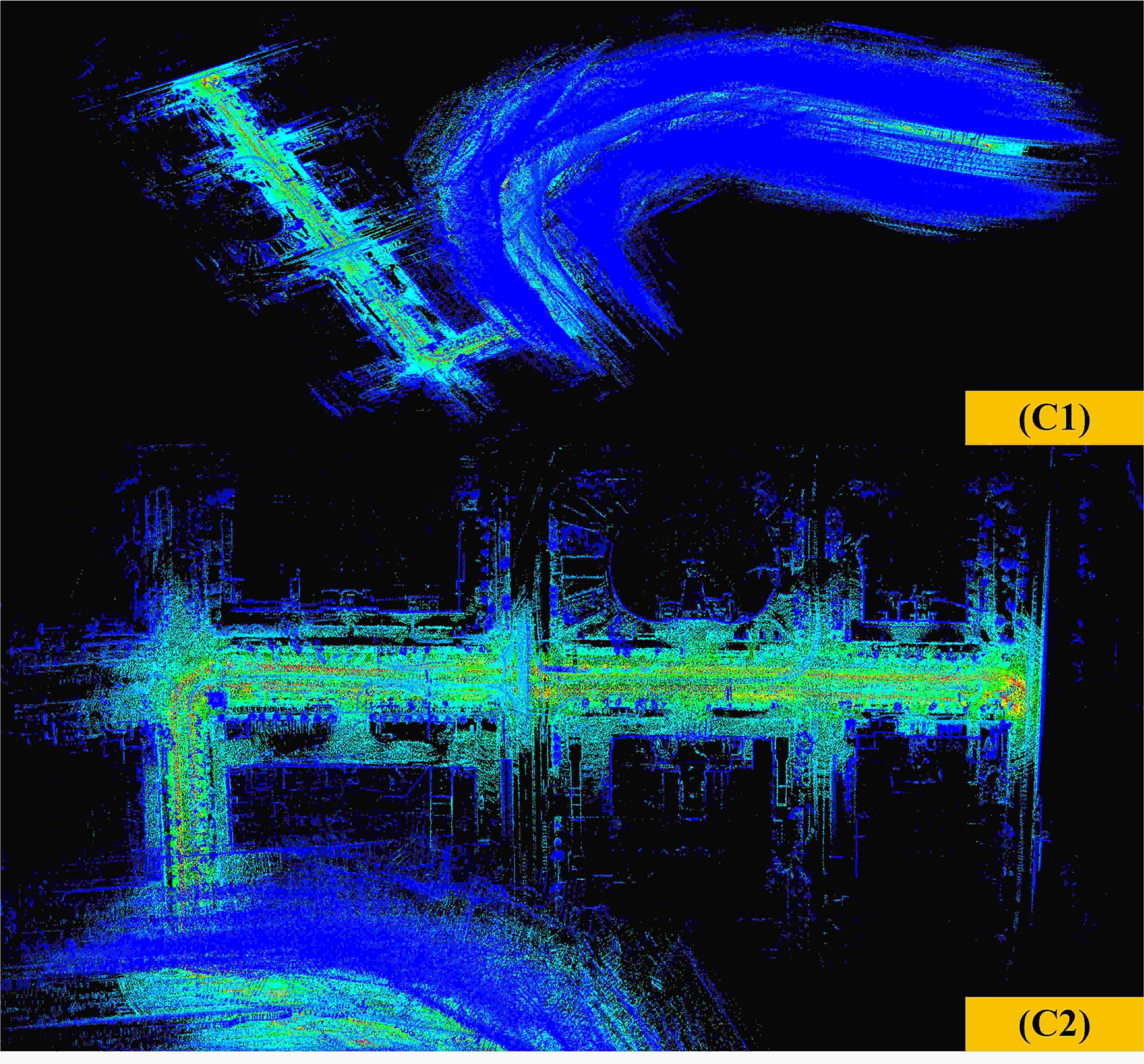}
	}
	
	\caption{The comparison of mapping after applying our, SuMa and LOAM algorithms respectively in dynamic highway environments. A2, B2 and C2 present the detail of mapping results for readers to compare the local registration accuracy of different methods.}
	\label{fig:dynamic}
\end{figure*}
\begin{figure*}[h]
	\centering
	\includegraphics[width=2.0\columnwidth]{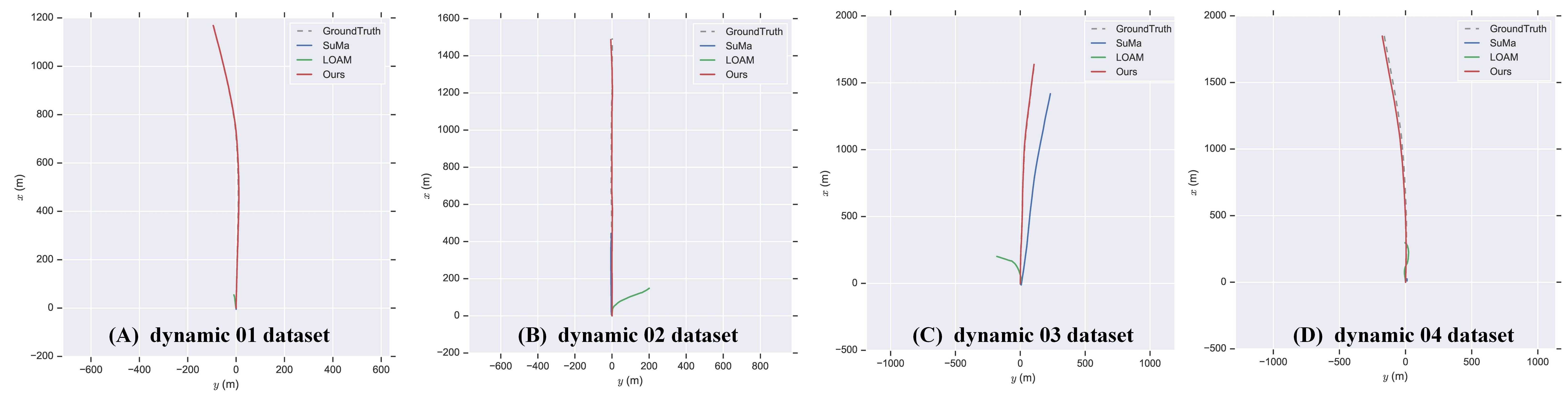}
	\caption{The comparison of estimated trajectories after applying LOAM, SuMa and Ours  algorithms in dynamic highway environments. The estimated trajectory from our method (red line) can align ground truth (dashed line) with high accuracy. In contrast, it is difficult for LOAM(green) and SuMa(blue).  }
	\label{fig:tencent_path}
	\vspace{-3mm}
\end{figure*}

To evaluate the robustness of ours, LOAM and SuMa algorithm, we use the raw data of laser measurement and compare the quality of the constructed map. The results are shown in Fig.\ref{fig:dynamic}(a), Fig.\ref{fig:dynamic}(b) and Fig.\ref{fig:dynamic}(c) respectively.  In Fig.\ref{fig:dynamic}(A1), we show the highway map built by our method and the yellow circle indicates the detail of map as shown in Fig.\ref{fig:dynamic} (A2). The red trajectory in Fig.\ref{fig:dynamic}(A2) represents the estimated motion of the vehicle. Along the red trajectory, we can see that the car moves from highway to urban, and the detail of map is always very clear. This shows that our method is very robust and can overcome the influence of moving objects. However, the highway map built by SuMa algorithm is not very good and the point cloud distortion marked with a blue circle is shown in Fig.\ref{fig:dynamic}(B2). This is because that SuMa cannot distinguish between static and moving objects. Therefore, the moving objects may cause wrong surfels in successive laser scans and affect the robustness of pose estimation.
Fig.\ref{fig:dynamic}(C1,C2) shows the map built by LOAM algorithm. We found that
the map is blurred, which shows that the LOAM algorithm also cannot overcome the influence of moving objects. Also, since the vehicle is driven from highway to urban areas, the number of geometric features varies significantly. In Fig.\ref{fig:dynamic}(C1,C2), the map built in an urban area is much better than the map in highway, which indicates that the LOAM algorithm relies heavily on geometric features. We also compare the performance of pose estimation on other dynamic datasets(01-04). During the test, we use Absolute Trajectory Error (ATE)\cite{sturm12iros} to evaluate the accuracy of each method and all sequences are 6-DoF aligned with the ground truth. The qualitative and quantitative results are shown in Fig.\ref{fig:tencent_path} and Table \ref{tab:accuracy}. The results indicate that our method performs better than state-of-the-art methods, LOAM and SuMa, in real highway environments. 


\begin{table}[h]
	\begin{center}
		\caption{Accuracy evaluation of LOAM SuMa and ours method on highway datasets (The blue numbers highlight the best result for each specific dataset)}
		\setlength{\tabcolsep}{1pt}
		\newcommand{\tabincell}[2]{\begin{tabular}{@{}#1@{}}#2\end{tabular}}
		\begin{tabular}{llllllllllllp{1cm}}
			\toprule
			\multirow{2}{*}{} 						& \multicolumn{3}{c}{ATE (in m) Transl. RMSE} 	& \multicolumn{4}{c}{ATE (in m) Transl. MAX}    	   \\ \midrule
			Datasets  					  			& LOAM     	& SuMa     	& OURS          & LOAM     	& SuMa      	& OURS      	& Length(m)     \\ \midrule
			dynamic 01                				& fail   	& fail    & \textcolor{blue}{2.186}  & fail     	& fail    	&\textcolor{blue}{4.552}    	& 1173   \\
			dynamic 02  					& fail   	& fail   & \textcolor{blue}{2.75}   & fail   &fail       & \textcolor{blue}{8.518} 	& 1491\\
			dynamic 03  					& fail   	& 45.97   & \textcolor{blue}{2.05}   & fail   &186.6       & \textcolor{blue}{4.59} 	& 1642\\
			dynamic 04  					& fail   	& fail   & \textcolor{blue}{2.9}   & fail   &fail       & \textcolor{blue}{15.93} 	& 1874\\
			dynamic 05  					& fail   	& 33.06   & \textcolor{blue}{4.94}   & fail   &72.29       & \textcolor{blue}{12.35} 	& 4580\\
			static  06  					& 23.1   	& 81   & \textcolor{blue}{12.4}   & 68.7   &213.4       & \textcolor{blue}{38.9} 	&6324\\
			\midrule
		\end{tabular}
		\label{tab:accuracy}
		\vspace{-8mm}
	\end{center}
\end{table}

\subsection{The comparison of pose estimation on KITTI datasets} 
\begin{figure}[H]
	\centering
	\includegraphics[width=0.9\columnwidth]{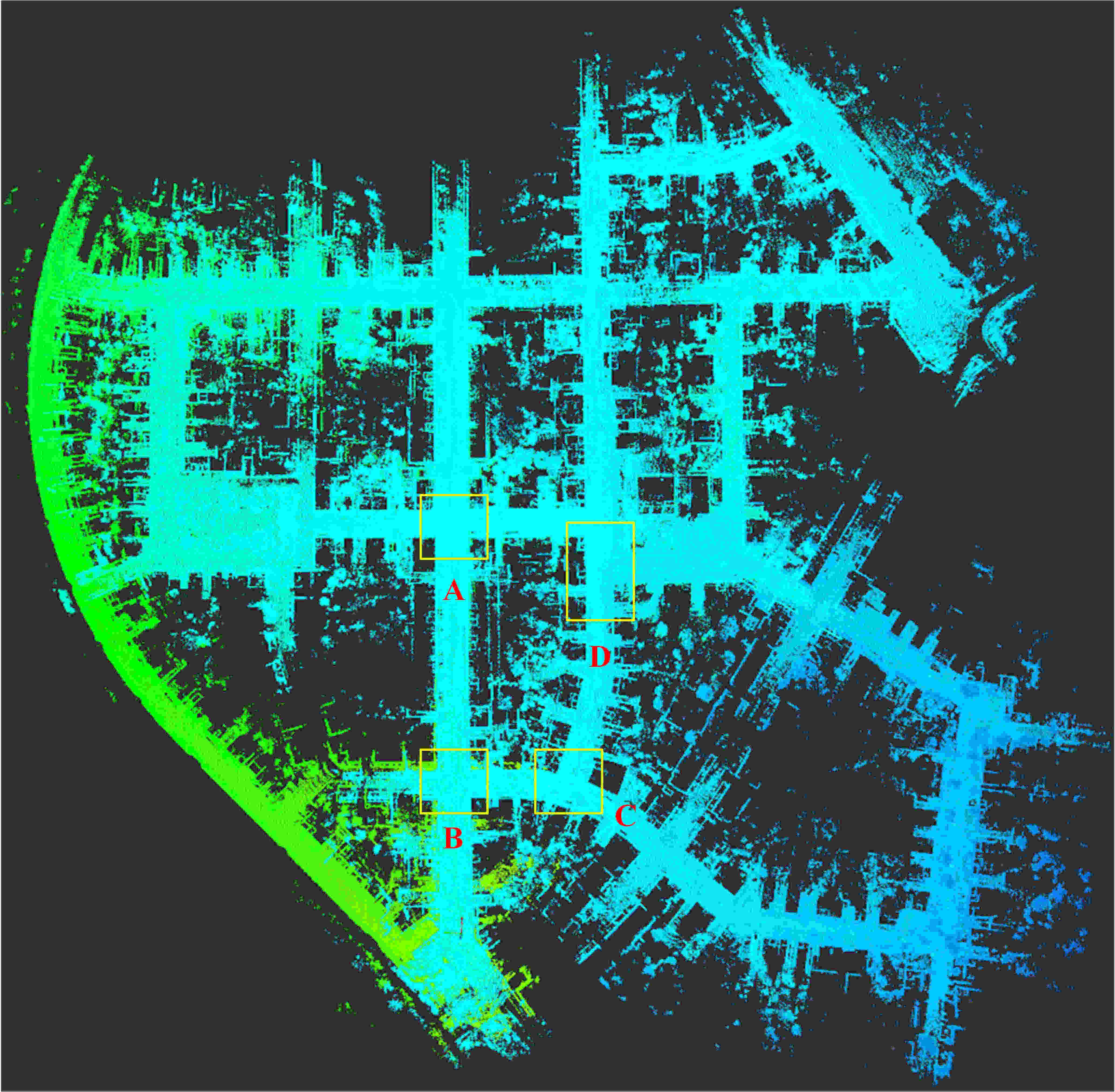}
	\caption{The performance of mapping results after applying Ours in KITTI 00 sequence. The color of map reflects the change in elevation. }
	\label{fig:kitti}
	\vspace{-3mm}
\end{figure}

In this section, we evaluate our approach on KITTI datasets which contain various environments, including urban, highway, etc. Fig. \ref{fig:kitti} presents the map built by our method on the KITTI 00 sequence. Some loop closure areas (A)-(D) are marked with red squares. Carefully examining these loop closure areas, we can see that there is only small point cloud distortion in the B area, which shows that our method can also achieve accurate pose estimation in urban environments. Note that in this work, our goal is to achieve accurate and robust laser odometry. Hence no loop closure is performed.

\begin{figure*}[h]
	\centering
	\includegraphics[width=1.7\columnwidth,height=1.1\columnwidth]{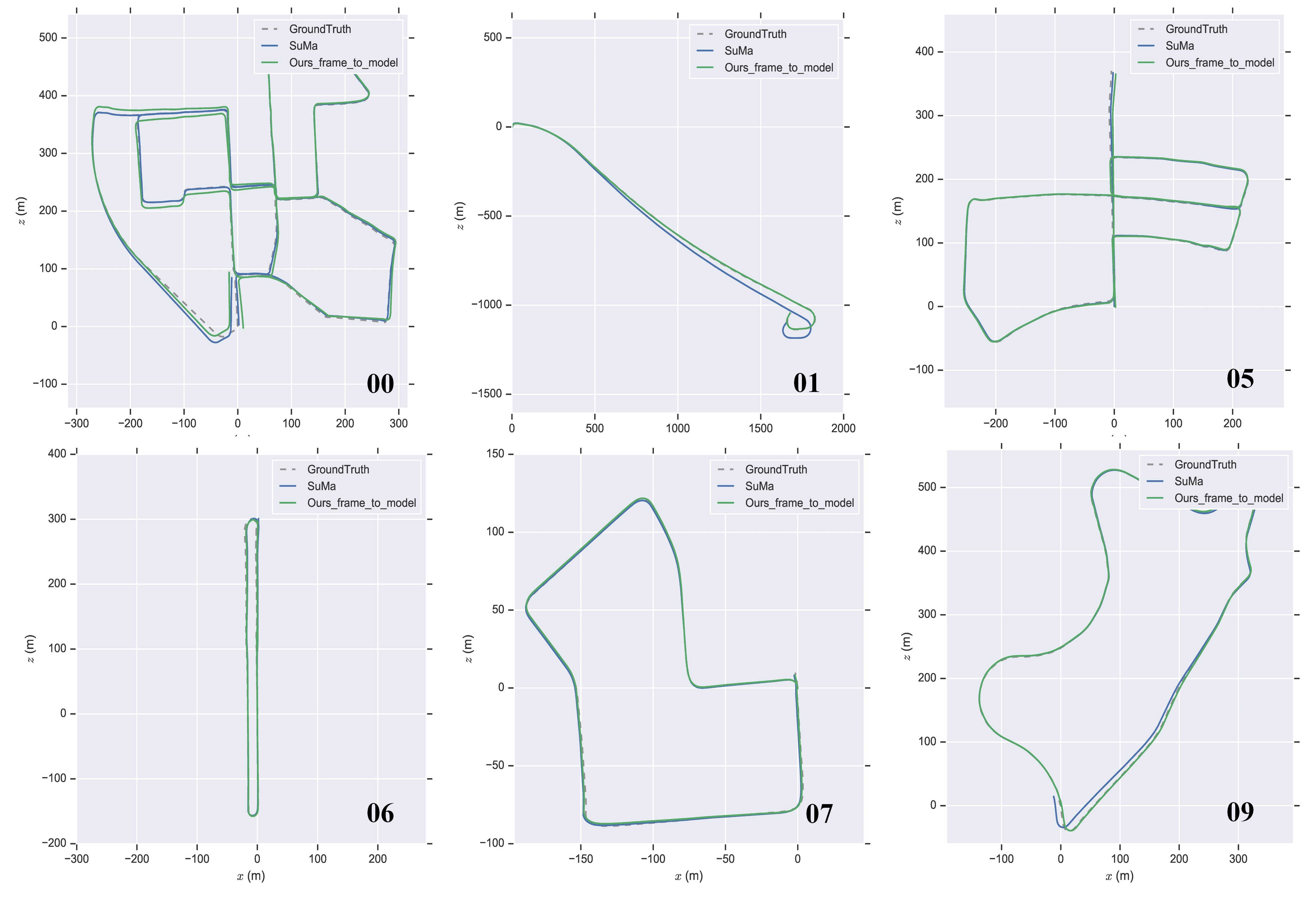}
	\caption{The comparison of estimated trajectories after applying Ours and SuMa algorithms on KITTI dataset.}
	\label{fig:path2}
\end{figure*}

Fig.\ref{fig:path2} presents the intuitive trajectory comparison between ours and SuMa algorithm on KITTI datasets. Table \ref{tab:kitti} lists the accuracy comparison of pose estimation on the proposed method without using the map (frame-to-frame), using the map (frame-to-model), LOAM and SuMa algorithm. It can be seen that our method achieves average translation error $0.85\%$ in comparison with $0.81\%$ and $0.88\%$ translation error of LOAM and SuMa respectively. This indicates that our method can achieve state-of-the-art performance. Especially on KITTI 01 sequence (highway environment), the relative translation error of the proposed algorithm is only $1.0\%$, in comparison with $1.4\%$ and $1.7\%$ of LOAM and SuMa algorithm respectively. This shows that the proposed algorithm is more robust compared to the other two methods in highway environments.

\begin{table*}[htbp]
	\begin{center}
		\caption{Accuracy evaluations of LOAM, SuMa and our method on KITTI datasets (The blue numbers highlight the best result for each specific dataset among all the compared algorithms)}
		\setlength{\tabcolsep}{10pt}
		\newcommand{\tabincell}[2]{\begin{tabular}{@{}#1@{}}#2\end{tabular}}
		\begin{tabular}{llllllllllllp{1cm}}
			\toprule
			\multirow{2}{*}{} 	& \multicolumn{10}{c}{Relative errors averaged over trajectories of 100m to 800m length: relative translation error in $\%$} \\ \midrule
			Approach  					  			& 00     	& 01     	& 03     & 04     	& 05      	& 06     &07  &09  &10 	& Average  \\ \midrule
			OURS(Frame-to-Frame)                	& 1.3   	& 3.4       & 1.4    & 1.2     	& 1.6    	&1.3     &2.1 &2.3 &1.2 &1.75 \\
			OURS(Frame-to-Model)  					& 0.9   	& \textcolor{blue}{1.0}   &\textcolor{blue}{0.9}   & \textcolor{blue}{0.6} &0.9 &0.6 &1.2   &\textcolor{blue}{0.8}       & \textcolor{blue}{0.8} 	& 0.85 \\
			LOAM \cite{zhang2017low}  					& 0.8   	& 1.4   & \textcolor{blue}{0.9}   & 0.7   &\textcolor{blue}{0.6}       & 0.7 &\textcolor{blue}{0.6} &\textcolor{blue}{0.8} & \textcolor{blue}{0.8}& \textcolor{blue}{0.81}\\
			SuMa (Frame-to-Model)				& \textcolor{blue}{0.7}   	& 1.7   & 1.0   & \textcolor{blue}{0.6}   &\textcolor{blue}{0.6}       &\textcolor{blue}{0.5} 	& 0.7 &0.9 &0.9 &0.88\\
			\midrule
		\end{tabular}
		\label{tab:kitti}
	\end{center}
\end{table*}

\subsection{Runtime Performance Evaluation}
In this section, we will present the runtime performance of each module. We test our algorithm with a 3.2-GHz i7 Intel processor and an integrated GPU (Nvidia GeForce GTX 1050 Ti), using the Robot Operating System (ROS) in Ubuntu Linux. For the experiment in Fig \ref{fig:dynamic}, we use a voxelized grid approach with a leaf size 0.5 to downsample the point cloud for each laser scan, and the runtime performance is shown in Table \ref{tab:runtime}. Note that all the listed modules will run on different threads. With the help of IMU measurement, the
	laser-inertial odometry runs at a frequency around 100 Hz, and its output will be further refined by laser mapping, which matches the undistorted cloud with a global map at a frequency of 1Hz.

\begin{table}[H]
	\begin{center}
		\caption{Runtime Performance for each module}
		\setlength{\tabcolsep}{2pt}
		\newcommand{\tabincell}[2]{\begin{tabular}{@{}#1@{}}#2\end{tabular}}
		\begin{tabular}{p{4cm} p{2cm}}
			\toprule
			\multirow{1}{*}{} 					
			Modules  					  			& Average Time \\ \midrule
			Scan Pre-processing                 	& 5ms       \\
			Dynamic Object Detection  				& 45ms    \\
			Laser-Inertial Odometry  				& 10ms    \\
			Laser Mapping   					    & 90ms    \\
			\midrule
		\end{tabular}
		\label{tab:runtime}
		\vspace{-8mm}
	\end{center}
\end{table}
\vspace{-1mm}
\section{Conclusion} 
\label{sec:conclusion}
In this paper, we present a robust laser-inertial and mapping method for highway environments, which utilizes both CNN segmentation network and laser-inertial framework and achieves real-time, low drift and robust pose estimation for large-scale highway environments. A large number of qualitative and quantitative experiments show that our method performs better than state-of-the-art methods in highway environment and achieves similar or better accuracy on KITTI dataset. In the future, we will try to achieve sensor fusion using laser, camera and IMU to further improve the robustness of the algorithm.
\section*{ACKNOWLEDGMENT}
 This work was supported by the National Natural Science Foundation of China (No. 61573091), Natural Science Foundation of Liaoning Province (20180520006) and Fundamental Research Funds for the Central Universities (No. N182608003, N172608005).
 We would also like to thank the Tencent Autonomous Driving Lab for providing the real highway dataset and allowing us to evaluate our method. We also wish to thank Dr.Ma Yanhai for his constructive advice.  

\balance

\bibliographystyle{IEEEtran}
\bibliography{mybibfile}



\end{document}